%% file: main.tex
\pdfoutput=1
\documentclass[11pt]{article}

\usepackage[nohyperref]{acl}
\usepackage{multirow}
\usepackage{times}
\usepackage{latexsym}
\usepackage{amsfonts}
\usepackage[T1]{fontenc}
\usepackage{booktabs}
\usepackage{colortbl}
\usepackage{multirow}
\usepackage{graphicx}
\usepackage{rotating}
\usepackage{placeins}

\usepackage{twemojis}
\usepackage[utf8]{inputenc}

\usepackage{microtype}

\usepackage{inconsolata}

\usepackage{packages/colab}
\usepackage{amsmath}
\usepackage{algorithm}
\usepackage[noend]{algpseudocode}

\title{Language Guided Exploration for RL Agents in Text Environments}

\author{Hitesh Golchha$^{\bullet}$, Sahil Yerawar$^{\bullet}$, Dhruvesh Patel$^{\bullet}$,
Soham Dan$^{\triangle}$, Keerthiram Murugesan $^{\triangle}$ \\ 
$^{\bullet}$Manning College of Information \& Computer Sciences,  University of Massachusetts Amherst  \\~~~~~ $^{\triangle}$IBM Research \\
\texttt{\{hgolchha,syerawar,dhruveshpate\}@cs.umass.edu}\\ \texttt{\{soham.dan,keerthiram.murugesan\}@ibm.com}}

\begin{document}
\maketitle

\input{sections/00_abstract_short}
\input{sections/01_introduction}

\input{sections/02_related_work}

\input{sections/03_method}
\input{sections/04_results}

\input{sections/05_conclusion}
\FloatBarrier

\section{Limitations}
Our work is the first to use a pre-trained language model as a guide for RL agents in text environments.
This paper focuses on the ScienceWorld environment, which is an English only environment. Moreover, it focuses mainly on scientific concepts and skills. 
To explore other environments in different languages with different RL agents will be an interesting future work. 
\bibliography{main}
\clearpage
\newpage
\input{sections/100_appendix}



\end{document}

%% file: sections/00_abstract_short.tex
\begin{abstract}
Real-world sequential decision making is characterized by sparse rewards and large decision spaces, posing significant difficulty for experiential learning systems like \textit{tabula rasa} reinforcement learning (RL) agents. Large Language Models (LLMs), with a wealth of world knowledge, can help RL agents learn quickly and adapt to distribution shifts. In this work, we introduce Language Guided Exploration (LGE) framework, which uses a pre-trained language model (called \textsc{Guide}) to provide decision-level guidance to an RL agent (called \textsc{Explorer}\twemoji[height=1em]{telescope}). We observe that on ScienceWorld \cite{Wang2022ScienceWorldIY}, a challenging text environment, LGE outperforms vanilla RL agents significantly and also outperforms other sophisticated methods like Behaviour Cloning and Text Decision Transformer.

\end{abstract}

%% file: sections/01_introduction.tex
\section{Introduction}\label{sec:introduction}

Reinforcement Learning (RL) has been used with great success for sequential decision making tasks. 
AI assistants whether text based \citep{li2022pre,huang2022language} or multi-modal \cite{chang2020procedure, vlamp}, have to work with large action spaces and sparse rewards. 
In such settings, the approach of random exploration is inadequate.
One needs to look for ways to use external information either to create a dense reward model or to reduce the size of action space.
In this work we focus on the latter approach.
\begin{figure}
    \centering
    \includegraphics[width=\columnwidth]{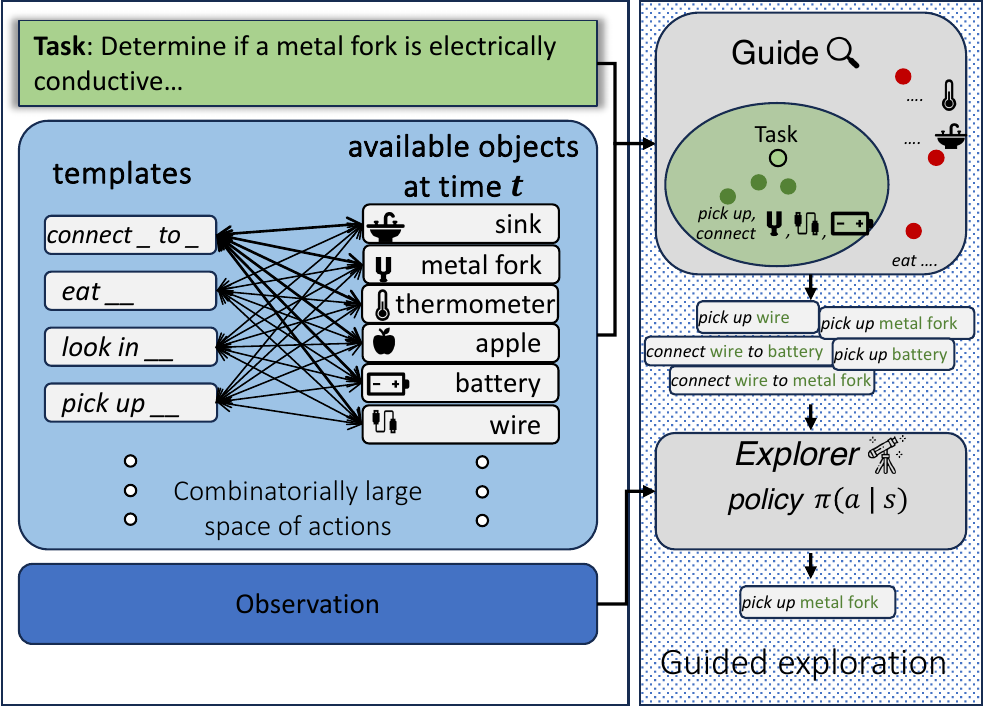}
    \caption{The Language Guided Exploration (LGE) Framework: The \emph{Guide} uses contrastive learning to produce a set of feasible action given the task description thereby reducing substantially the space of possible actions. The \emph{Explorer}, an RL agent, then uses the set of actions provided by the \emph{Guide} to learn a policy and pick a suitable action using it.
    }
    \label{fig:framework}
\end{figure}

We make a simple observation that, in many cases, the textual description of the task or goal contains enough information to completely rule out certain actions, thereby greatly reducing the size of the effective action space.
For example, as shown in~Fig.\ref{fig:framework}, if the task description is \emph{``Determine if a metal fork is electrically conductive''}, then one can safely rule out actions that involve objects like sink, apple, and actions like eat, smell, etc.
Motivated by this observation, we introduce the \textbf{L}anguage \textbf{G}uided \textbf{E}xploration (LGE) framework that uses an RL agent but augments it with a \emph{Guide} model that uses world knowledge to rule out large number of actions that are infeasible or highly unlikely. 
Along with removing irrelevant actions, the frameworks supports generalization in unseen environments where new objects may appear. For example, if the model observed an apple in the environment during training, at test time, the environment may contain an orange instead. But the guide, which posses commonsense may understand that all fruits are equally relevant or irrelevant for the given task.

To test our framework, we use the highly challenging benchmark called \textsc{ScienceWorld} \citep{Wang2022ScienceWorldIY}, which consists of a purely text based environment where the observations, actions, and inventory are expressed using natural language text.
\textsc{ScienceWorld} embodies the major challenges faced by RL agents in realy world applications: the template based actions with slots for verbs and objects produce a combinatorially large action space, the long natural language based observations make for a challenging state representation, and the rewards signals based mainly on the completion of challenging tasks create a delayed and sparse reward signal. Following are the main contributions of our work:

\noindent ~ We propose a novel way to allow language guided exploration for RL agents. The task instructions are used to identify relevant actions using a contrastively trained LM. 
The proposed \textsc{Guide} model that uses contrastive learning has not been explored for text environments before. 

\noindent  We demonstrate significantly stronger results on the \textsc{ScienceWorld} environment when compared to methods that use Reinforcement Learning, and more sophisticated methods like Behaviour Cloning \citep{wang-etal-2023-behavior} and Text Decision Transformer \citep{chen2021decisiontransformer}.

%% file: sections/02_related_work.tex
\section{Related Work}\label{sec:related work}

Text-based environments \cite{zork1658697,yin2019learn,murugesan2020textbased,textworld2019} provide a low-cost alternative to complex 2D/3D environments, and real world scenarios, for the development of the high-level learning and navigation capabilities of the AI agents. 
Due to the complexity of these environments, \textit{tabula rasa} RL agents \citep{he-etal-2016-deep-reinforcement,zahavy10.5555/3327144.3327274,yao-etal-2020-keep} struggle to learn anything useful.
Therefore several methods like imitation learning, use of knowledge graphs \citep{Ammanabrolu2020Graph}, Case-Based Reasoning \cite{atzeni2022casebased}, behavior cloning \citep{chen2021decisiontransformer}, intrinsically motivated RL, and language motivated RL \citep{pmlr-v202-du23f, Adeniji2023LanguageRM} have been proposed. 
The main aim of all these methods is to use external knowledge or a handful of gold trajectories to guide the learning. 
In our work, we address the same issue in a much direct and generalizable manner by reducing the size of the action space using an auxiliary model called the Guide.

%% file: sections/03_method.tex
\section{Methodology}\label{sec:method}

\noindent\textbf{Notation: } The text environment, a partially observable Markov decision process (POMDP) consists of $(S, T, A, R, \tilde O, \Omega)$. 
In \textsc{ScienceWorld}, along with the description of the current state, the observation also consists of a task description $\tau\in\mathcal T$ that stays fixed throughout the evolution of a single trajectory, i.e., $\tilde O=  O\times \mathcal T$, where $O$ is the set of textual descriptions of the state and $\mathcal T$ is the set of tasks (including different variations of each task).
Note that the set of tasks are divided into different types and each type of task has different variations, i.e., $\mathcal T = \bigcup_{\gamma=1}^\Gamma \bigcup_{v=1}^{V_\gamma} \tau_{\gamma, v}$, where $\Gamma$ is the number of task types and $V_\gamma$ is the number of variations for the task type $\gamma$. 
Gold trajectories $G_{\gamma, v} = \{a_1, a_2, .. , a_T\}$ are available for each $\gamma$, $v$. 
 
\subsection{The LGE framework} \label{sec:the LGE framework}
We propose a Language Guided Exploration Framework (LGE), which consists of an an RL agent called the \textsc{Explorer} \  , and an auxiliary model that scores each action called the \textsc{Guide} \ . 
The \textsc{Explorer} starts in some state sampled from initial state distribution $d_{0}$. 
At any time step $t$, a set of all valid actions $A_{\gamma, v, t}$ is provided by the environment. 
This set, constructed using the cross product of action templates and the set of objects (see~Fig.\ref{fig:framework}) is extremely large, typically in thousands.
The \textsc{Guide} uses the task description $\tau_{\gamma, v}$, to produce a set of most relevant actions $\hat A_{\gamma, v, t} \subset A_{\gamma, v, t}$.
With a probability $1 - \epsilon$ ~(resp. $\epsilon$), the \textsc{Explorer} samples an action from ${\hat A}_{\gamma, v, t}$ using its policy $\pi(a|s_t)$~(resp., from $A_{\gamma, v, t}$).   
Algorithm \ref{alg:lge_algorithm} in Appendix \ref{app: implementation details} outlines the steps involved in the \textsc{LGE} framework using  a DRRN \cite{he-etal-2016-deep-reinforcement} based \textsc{Explorer}.

\subsubsection{\textsc{Explorer} \ }\label{sec:explorer}
The \textsc{Explorer} learns a separate policy $\pi_\gamma$ for each task type $\gamma\in \Gamma$ by exploring the the environment.\footnote{The agent learns a separate policy of each task type but this policy is common across all variations for that particular task type.}
We use the Deep Reinforcement Relevance Network (DRRN) \cite{he-etal-2016-deep-reinforcement} as our \textsc{Explorer}, as it has shown to be the strongest baseline in \citet{Wang2022ScienceWorldIY}. However, our framework allows to swap the DRRN with any other RL agent.
The DRRN uses Q-learning with with prioritized experience replay to perform policy improvement using a parametric approximation of the action value function $Q(s,a)$.\footnote{We follow the implementation of DRRN provided in \citet{Hausknecht2019InteractiveFG}.}
The current state $s_t$ is represented by concatenating the representations of the past observation $o_{t-1}$, 
inventory $i_t$ and look around $l_t$, each encoded by separate GRUs, i.e.,
$h_{s_t} = \left(\,f_{\theta_o}(o_{t-1})\,:\, f_{\theta_i}(i_t)\,:\, f_{\theta_l}(l_t)\,\right).$ 
Each relevant action $a \in A_{\text{rel},t}$ is encoded in the same manner:
$h_{a_t} = f_{\theta_a}(a_t).$ 
Here $f_*$ are the respective GRU encoders, $\theta_*$ their parameters and ``$\,:\,$'' denotes concatenation. 
The value function $Q(s,a)$ is represented using a linear layer over the concatenation of the action and state representations
$Q(s_t, a_t | \theta) = W^T \cdot \left( h_{s_t} : h_{a_t}\right) + b,$ where $\theta$ is a collection of $\theta_o$, $\theta_i$, $\theta_l$, $\theta_a$, $W$ and $b$.
During training, a stochastic policy based on the value function is used: $\hat a \sim \pi(a|s)\propto Q(s,a| \theta)$, while at inference time we use greedy sampling: $\hat a = \arg\max_{a} Q(s,a| \theta)$.    

\subsubsection{\textsc{Guide} \ }
While LLMs are capable of scoring the relevant actions without any finetuning, we observed that due to the idiosyncrasies of the \textsc{ScienceWorld} environment, it is beneficial to perform some finetuning.
We use SimCSE \citep{gao-etal-2021-simcse}, a contrastive learning framework, to finetune the \textsc{Guide} LM.  
The training data $\{\tau_i, G_i\}_{i=1}^M$, which consists of task descriptions $\tau_i = \tau_{\gamma, v} \in\mathcal T$ along with the set of corresponding gold actions $G_i=G_{\gamma, v}$. 
 The \textsc{Guide} model $g_\phi$ is used to embed the actions and the task descriptions into a shared representation space where the similarity score of a task and an action is expressed as
 $\mathop{s}(\tau, a) = \frac{g_{\phi}(\tau)\,\cdot\,g_{\phi}(a)}{\lambda}$, with $\lambda$ being the temperature parameter.
 The training objective is such that the embeddings of a task are close to those of the corresponding relevant actions, expressed using the following loss function:
 \begin{align*}
    l(\phi; \tau_i, G_i) = -\log \frac{e^{\mathop{s}(\tau_i, \,a^+)}}{e^{s(\tau_i, a^+)} + \sum\limits_{a^-\in  N_i} e^{s(\tau, a^-)} },
 \end{align*}
 where $a^+\sim G_i$ is a relevant action and $N_i$ is a fixed sized subset of irrelevant actions.\footnote{Details of the models used and the training data are provided in Appendix \ref{app: implementation details}.}
%

Note that since we only have access to a small amount of gold trajectories (3442) for training, we take special steps to avoid overfitting, which is the main issue plaguing the imitation learning based methods. 
First, we only provide the task description to the \textsc{Guide} and not the full state information. Second, unlike the \textsc{Explorer}, which uses different policy for each task type,  we train a common \textsc{Guide} across all tasks.

%% file: sections/04_results.tex
\section{Experiments and Results}\label{sec:results}

As done in \citet{Wang2022ScienceWorldIY}, the variations of each task type are divided into training, validation and test sets.
Both \textsc{Guide} and \textsc{Explorer} are trained only using the training variations.

\input{tables/evaluating_guide}

\subsection{Evaluating the \textsc{Guide}}
Before the joint evaluation, we evaluate the \textsc{Guide} in isolation.
We sample 5 variations from the validation set for each task type and compute the three metrics: GAR, RST and MAP.
We use the following two intuitive but strong baselines: 

\noindent\textbf {(1) Gold per-task ($G_\tau$)}: We create a set of 50 most most used actions in gold trajectories of all training variations of a particular task. The Gold per-task baseline, predicts an action to be relevant if it belongs to this set.

\noindent\textbf {(2) Gold Global ($G_g$) }:  Similar to Gold per-task but we use 50 most used actions in Gold trajectories of all training variations for all tasks.


\paragraph{Gold Action Rank (GAR):} 
At any time step $t$, $\mathop{GAR}(\gamma, v, t)$ is defined as the rank of the gold $a_t$ in the set of valid actions $A_{\gamma, v, t}$, and the Gold Action Reciprocal Rank (GARR) is defined as 1/GAR.
Since the size of $A_{\gamma, v, t}$ is variable for every $t$, we also report percent GAR. 
As seen in Table \ref{tab:Evaluating Guide}, {the gold action gets an average rank of $7.42$, which is impressive because $|A_{\gamma, v, t}|$ averages around 2000.}

\paragraph{Relevant Set Recall (RSR):}  
GAR ranks a single optimal action at any time, but multiple valid action sequences may exist for task completion. Although all viable paths are not directly accessible, we estimate them. For each time step $t$ in variation $\tau_{\gamma, v}$, a set of gold relevant actions $\tilde A_{\gamma, v, t}$ is identified by intersecting the gold trajectory $G_{\gamma, v}$ with valid actions at $t$, so $\tilde A_{\gamma, v, t}= \{a\,|\, a\in G_{\gamma, v} \cap A_{\gamma, v, t}\}$. The \textsc{Guide}'s effectiveness is measured by its recall of this set, considering its top-k predicted actions $\hat A_{\gamma, v, t}$. Relevant Set Recall (RSR) is calculated as $RSR(\gamma, v, t) = \frac{|\hat A_{\gamma, v, t}\cap \tilde A_{\gamma, v, t}|}{|\tilde A_{\gamma, v, t}|}.$ 
\noindent As seen in Table \ref{tab:Evaluating Guide}, {the \textsc{Guide} has almost perfect average recall of 0.99} while selecting top 50 actions for the \textsc{Explorer} at every step of the episode. 
\input{tables/ex1}
\paragraph{Mean Avg. Precision (MAP):} 
The \textsc{Guide} also functions as a binary classifier, predicting the relevance of each action in $A_{\gamma, v, t}$. Using a threshold-free metric like average precision score \citep{scikit-learn}, the \textsc{Guide} achieves a superior average precision score of 0.68 compared to baselines. 
Coupled with perfect recall at 50, this indicates the \textsc{Guide}'s strong generalization ability on new variations and robust performance across various thresholds.
We observe that the threshold that produces best MAP is 0.52, which corresponds to ${|\hat A_{\gamma, v, t}|}=28$ on average. So, to be conservative, we use $k=50$ in the subsequent evaluations. Table \ref{tab:qualitative-analysis_valid_trajectories_task0} shows an example of the set of actions selected by \textsc{Guide} for the task ``Change of state''.

\subsection{Evaluating LGE}
We follow the same evaluation protocol as \citep{Wang2022ScienceWorldIY} and evaluate two versions of the LGE framework, one with a fixed $\epsilon$ of 0.1 and the other with $\epsilon$ increasing from 0 to 1.
Table \ref{tab:agent-performance} reports the means returns for each task.

\noindent\textbf{LGE improves significantly over the RL baseline.} 
The DRRN agent, which only uses RL, performs the best among the baselines.
The proposed LGE framework (last two columns), improves the performance of DRRN on 18 out of 30 tasks. On average the LGE with $\epsilon=0.1$, improves the mean returns by $35\%$ ($0.17 \to 0.23$).

\noindent\textbf{LGE is better than much more complex, specialized methods.} 
The behaviour cloning (BC) model, uses a Macaw \citep{Tafjord2021Macaw} model finetuned on the gold trajectories to predict the next action. The Text Decision Transformer (TDT) \citep{chen2021decisiontransformer} models the complete POMDP trajectories as a sequence and is capable of predicting actions that maximize long-term reward. 
As seen in Table \ref{tab:agent-performance}, the simpler LGE framework outperforms both TDT and BC. This shows the importance of having an RL agent in the framework that can adapt to the peculiarities of the environment.


\noindent\textbf{Increasing $\epsilon$ does not always help.}
$\epsilon=1$ corresponds using only the \textsc{Explorer}—ideal once the policy is trained well. 
However, we observe that the actions provided by the \textsc{Guide} almost always contain the right action and increasing $\epsilon$ does not always help.


\input{tables/main_table}

%% file: tables/evaluating_guide.tex
\begin{table}[]
\centering
\resizebox{\columnwidth}{!}{%
\begin{tabular}{@{}lcccccc@{}}
\toprule
\textbf{Model} &
  \textbf{top-k} &
  \textbf{RSR} &
  \textbf{MAP} &
  \textbf{GAR} &
  \textbf{GAR (\%)} &
  \textbf{GARR} \\ \midrule
$G_g$ &
  50 &
  0.71 &
  0.52 &
  \multirow{2}{*}{N/A} &
  \multirow{2}{*}{N/A} &
  \multirow{2}{*}{N/A} \\
$G_\tau$ & 50 & 0.9  & 0.66 &  &  &  \\ \midrule
\multirow{3}{*}{Guide} &
  50 &
  0.99 &
  0.68 &
  \multirow{3}{*}{7.4  \small{± 16.2}} &
  \multirow{3}{*}{1.8 \small{± 9.4}} &
  \multirow{3}{*}{0.31 \small{± 2.3}} \\
              & 20 & 0.94 & 0.67 &  &  &  \\
              & 10 & 0.79 & 0.61 &  &  &  \\ \bottomrule
\end{tabular}%
}
\caption{Various metrics used to evaluate the \textsc{Guide} in isolation. Note that for the baselines $G_g$ and $G_\tau$, we cannot compute GAR.}
\label{tab:Evaluating Guide}
\end{table}

%% file: tables/ex1.tex
\begin{table*}[!h]
\centering
\resizebox{0.8\textwidth}{!}{%
\begin{tabular}{|l|l|}
    \toprule
    Relevant Gold Actions & Selected By \textsc{Guide}  \\
    \toprule
      open cupboard,focus on soap in kitchen, & \textcolor{darkgreen}{pick up metal pot},move metal pot to sink,pour metal pot into metal pot, \\
     
    \textcolor{red}{move soap in kitchen to metal pot}, & \textcolor{darkgreen}{open cupboard,
activate stove,move metal pot to stove,
pick up thermometer},
 \\ 
    move metal pot to stove, & open freezer,
wait,
\textcolor{darkgreen}{go to outside},
open glass jar,look around,
open drawer in cupboard, 
 \\
 go to outside,wait1, & open drawer in counter,open oven,move ceramic cup to sink,pick up ceramic cup,open fridge, \\
 pick up thermometer, &open door to hallway, activate sink, mix metal pot, pour ceramic cup into ceramic cup, \\
 pick up metal pot,look around,activate stove & pick up sodium chloride, \textcolor{darkgreen}{wait1}, focus on metal pot, pick up soap in kitchen
\\ 
\bottomrule
\end{tabular}
}
\caption{Column 1 shows the relevant gold actions for the task ``Change of State (variation 1 from the dev set)'', and column two shows the set of actions selected by the \textsc{Guide}. The missed gold actions are in \textcolor{red}{Red}, while selected gold actions are in \textcolor{darkgreen}{Green}}
\label{tab:qualitative-analysis_valid_trajectories_task26} 
\end{table*}

%% file: tables/main_table.tex
\begin{table}[!t]
\centering
\resizebox{0.9\columnwidth}{!}{%
\begin{tabular}{lcccccccc}
\toprule
 \textbf{\small Task} & DRRN*  & BC* & TDT* & LGE {\small inc} & LGE {\small fix} & Delta \\ 
\toprule
\rowcolor[HTML]{D3D3D3} 
            T0     & 0.03  & 0.00 & 0.00 & \textbf{0.04} & 0.02 & 0.01($\uparrow$) \\ 
\rowcolor[HTML]{D3D3D3} 
T1                    & 0.03  & 0.00 & 0.00 & 0.02 & 0.03  & 0.00 \\ 
\rowcolor[HTML]{D3D3D3} 
T2                & 0.01  & 0.01 & 0.00 & 0.00 & 0.00 & -0.01($\downarrow$) \\ 
\rowcolor[HTML]{D3D3D3} 
T3                   & \textbf{0.04}  & 0.00 & 0.01 & 0.02 & 0.03 & -0.01($\downarrow$) \\ 
\hline  

T4             & 0.08  & 0.01 & 0.02 & 0.08 & 0.08 & 0.00 \\ 
T5           & 0.06  & 0.01 & 0.02 & 0.06 & \textbf{0.07} & 0.01($\uparrow$) \\  
T6                             & 0.10  & 0.04 & 0.04 & 0.08 & \textbf{0.11} & 0.01($\uparrow$)  \\ 
\hline 
\rowcolor[HTML]{D3D3D3} 
T7                            & 0.13  & 0.03 & 0.07 & 0.13 & 0.13 & 0.00 \\ 
\rowcolor[HTML]{D3D3D3} 
T8           & \textbf{0.10}  & 0.02 & 0.05 & 0.08 & 0.1 & -0.02($\downarrow$) \\ 
\rowcolor[HTML]{D3D3D3} 
T9                   & \textbf{0.07}  & 0.05 & 0.05 & 0.06 & 0.06 & -0.01($\downarrow$) \\ 
\rowcolor[HTML]{D3D3D3} 
T10                 & 0.20  & 0.04 & 0.05 & 0.23 & \textbf{0.29} & 0.09($\uparrow$) \\ \hline 
T11                              & 0.19  & 0.21 & 0.19 & \textbf{0.39} & 0.19 & 0.20($\uparrow$) \\ 
T12                         & 0.26  & 0.29 & 0.16 & 0.18 & \textbf{0.56} & 0.30($\uparrow$) \\ 
T13                     & 0.56  & 0.19 & 0.17 & 0.55 & \textbf{0.60} & 0.04($\uparrow$) \\ 
T14                                 & 0.19  & 0.17 & 0.19 & 0.19 & \textbf{0.67} & 0.48($\uparrow$) \\ 

\hline 
\rowcolor[HTML]{D3D3D3} 
T15                                & 0.16  & 0.03 & 0.05 & \textbf{0.18} & 0.17 & 0.02($\uparrow$) \\ 
\rowcolor[HTML]{D3D3D3} 
T16                                & 0.09  & 0.08 & 0.03 & \textbf{0.10} & 0.094 & 0.01($\uparrow$) \\ 
\hline 
T17                            & 0.20  & 0.06 & 0.10 & 0.21 & \textbf{0.25} & 0.05($\uparrow$) \\ 
T18           & 0.29  & 0.16 & 0.20 & \textbf{0.30} & 0.27 & 0.01($\uparrow$) \\ 
T19            & 0.11  & 0.05 & 0.07 & 0.11 & 0.11 & 0.00 \\ \hline 
\rowcolor[HTML]{D3D3D3} 
T20               & 0.48  & 0.26 & 0.20 & 0.55 & \textbf{0.89} & 0.41($\uparrow$) \\ 
\rowcolor[HTML]{D3D3D3}
T21 & 0.31  & 0.02 & 0.20 & \textbf{0.33} & 0.32 & 0.02($\uparrow$) \\ 
\rowcolor[HTML]{D3D3D3} 
T22              & 0.47  & 0.14 & 0.16 & \textbf{0.64} & 0.46 & 0.17($\uparrow$) \\ 
\hline 
T23               & 0.10  & 0.02 & 0.07 & 0.17 & \textbf{0.18} & 0.08($\uparrow$) \\ 
T24                & 0.09  & 0.04 & 0.02 & \textbf{0.16} & 0.05 & 0.07($\uparrow$) \\ 
\hline
\rowcolor[HTML]{D3D3D3} 
T25           & 0.13  & 0.05 & 0.04 & 0.24 & \textbf{0.25} & 0.12($\uparrow$) \\ 
\rowcolor[HTML]{D3D3D3} 
T26                   & 0.13  & 0.05 & 0.04 & \textbf{0.25} & 0.24 & 0.12($\uparrow$) \\ 
\rowcolor[HTML]{D3D3D3} 
T27                 & 0.13  & 0.04 & 0.04 & 0.21 & 0.21 & 0.08($\uparrow$) \\ \hline 
T28           & 0.19  & 0.06 & 0.06 & 0.19 & \textbf{0.22} & 0.03($\uparrow$) \\ 
T29         & 0.17  & 0.13 & 0.05 & 0.17 & 0.16 & 0.00 \\ 
\midrule
     \multicolumn{1}{l}{\textbf{Avg.}}                                          &   {0.17}   &   {0.08}  &   {0.08} & {0.20} & \textbf{0.23} & 0.06($\uparrow$) \\
\bottomrule
\end{tabular}
}
\caption{Zero-shot performance of the agents on test variations of across all tasks. The columns with * are reported from \citet{Wang2022ScienceWorldIY}. The Delta column is the difference between DRRN and the best LGE model. { The names of the tasks are in Table \ref{tab:task-names} in Appendix.}}
\label{tab:agent-performance} 
\end{table}

%% file: sections/05_conclusion.tex
\section{Conclusion}\label{sec:conclusion}

We proposed a simple and effective framework for using the knowledge in LMs to guide RL agents in text environments, and showed its effectiveness on the \textsc{ScienceWorld} environment when used with DRRN. 
Our framework is generic and can extend to work with other RL agents. 
We believe that the positive results observed in our work will pave the way for future work in this area.   

%% file: sections/100_appendix.tex
\appendix

\section{Appendix}\label{sec:appendix}

\subsection{Implementation details}\label{app: implementation details}

\subsubsection{\textsc{Guide}'s architecture}
We use a BERT-base model \cite{Devlin2019BERTPO} as the \textsc{Guide}. 
We also performed a rudimentary experiment of fine-tuning the Encoder part of the 770M Macaw \cite{Tafjord2021Macaw} model (T5 Large model pretrained on Question Answering datasets in Science Domain),  but could not achieve the same quality of pruning post training as the smaller BERT-base model. This could be attributed to two reasons: 
\begin{enumerate}
    \item The size of the training dataset may not be enough to train the large number of parameters in the bigger Macaw model (thus leading to underfitting). 
    \item We used a smaller batch size for training the Macaw model using similar compute as the BERT-base model (16GB GPU memory). As the contrastive loss depends on in-batch examples for negative samples, the smaller batch-size could mean less effective signal to train the model. We would explore a fairer comparison with similar training settings as the BERT model in future work.  
\end{enumerate}

The  code for this work is available at at this repository: \url{https://github.com/hitzkrieg/drrn-scienceworld-clone}
\input{tables/task_names}

\subsubsection{Training the \textsc{Guide}}

The supervised contrastive loss framework in \cite{gao-etal-2021-simcse} needs a dataset consisting of example triplets of form  ($x_i$, $x_i^+$ and $x_i^-$) where $x_i$ and $x_i^+$ are semantically related and $x_i^-$ is an example of a hard negative (semantically unrelated to $x_i$, but more still more similar than any random sample). 

For training the Guide, we want to anchor the task descriptions closer in some embedding space to relevant actions and away from irrelevant actions. Thus we prepare a training data $\{(\tau_i, a_i^+, a_i^-)\}_{i=1}^M$,  consists of tuples of task descriptions $\tau_i = \tau_{\gamma, v} \in\mathcal T$ along with a relevant action $a_i^+ \sim G_{\gamma, v}$ and an irrelevant action $a_i^- \sim \mathcal{N}_{\gamma}$ (fixed size set of irrelevant actions for every task $\gamma$).

Preparing  $\mathcal{N}_{\gamma}$: We simulate gold trajectories from 10 random training variations for each task-type $\gamma \in \Gamma$, and keep taking a union of the valid actions at each time step to create a large union of valid actions for that task-type. $\mathcal{N}_{\gamma} = \bigcup_{v=1}^{10}\bigcup_{t} A_{\gamma,v,t}$. Now, this set is used for sampling hard negatives for a given task description. For a batch of size N, the loss is computed as:
\begin{align}
  l(\phi) = - \sum_{i=1}^N \log \frac{e^{\mathop{s}(\tau_i, \,a_i^+)}}{\sum_{j=1}^N e^{s(\tau_i, a_j^-)} + e^{s(\tau_i, a_j^+)}},
\end{align}

The final training dataset to train the \textsc{Guide} LM on 30 task-types consisting of 3442 training variations had 214535 tuples. The LM was trained with a batch size of 128, on 10 epochs and with a learning rate of 0.00005. 

\subsubsection{Training and evaluating the {Explorer}}
We use similar approach as \cite{Wang2022ScienceWorldIY} to train and evaluate the Explorer. The DRRN architecture is trained with embedding size and hidden size = 128, learning rate = 0.0001, memory size = 100k, priority fraction (for experience replay) = 0.5. The model is trained simultaneously on 8 environment threads at 100k steps per thread. Episodes are reset if they reach 100 steps, or success/failure state.

After every 1000 training steps, evaluation is performed on 10 randomly chosen test variations. The final numbers reported in table \ref{sec:results}  are the average score of last 10\% test step scores. 

\input{tables/algo_wide}
\subsection{More examples}

Table \ref{tab:qualitative-analysis_valid_trajectories_task26} shows an example of the out of the \textsc{Guide}.
\input{tables/ex2}


%% file: tables/task_names.tex
\begin{table}[!h]
\resizebox{\columnwidth}{!}{%
\begin{tabular}{|l|l|}
\toprule
TaskID & Task Name                                   \\
\toprule
T0     & Changes of State (Boiling)                  \\
T1     & Changes of State (Any)                      \\
T2     & Changes of State (Freezing)                 \\
T3     & Changes of State (Melting)                  \\
T4     & Measuring Boiling Point (known)             \\
T5     & Measuring Boiling Point (unknown)           \\
T6     & Use Thermometer                             \\
T7     & Create a circuit                            \\
T8     & Renewable vs Non-renewable Energy           \\
T9     & Test Conductivity (known)                   \\
T10    & Test Conductivity (unknown)                 \\
T11    & Find an animal                              \\
T12    & Find a living thing                         \\
T13    & Find a non-living thing                     \\
T14    & Find a plant                                \\
T15    & Grow a fruit                                \\
T16    & Grow a plant                                \\
T17    & Mixing (generic)                            \\
T18    & Mixing paints (secondary colours)           \\
T19    & Mixing paints (tertiary colours)            \\
T20    & Identify longest-lived animal               \\
T21    & Identify longest-then-shortest-lived animal \\
T22    & Identify shortest-lived animal              \\
T23    & Identify life stages (animal)               \\
T24    & Identify life stages (plant)                \\
T25    & Inclined Planes (determine angle)           \\
T26    & Task 26 Friction (known surfaces)           \\
T27    & Friction (unknown surfaces)                 \\
T28    & Mendelian Genetics (known plants)           \\
T29    & Mendelian Genetics (unknown plants)         \\
\hline
\end{tabular}
}
\caption{List of Task Names with their task ID's}
\label{tab:task-names}
\end{table}

%% file: tables/algo_wide.tex
\begin{algorithm*}[]
\centering
\caption{Training Algorithm: \textsc{Language Guided Exploration Framework}}
\label{alg:lge_algorithm}
\begin{algorithmic}
\State Initialize replay memory \(D\) to capacity \(C\)
\State Initialize Explorer's Q-network with random weights \(\theta\)
\State Initialize \(updateFrequency\), \(totalSteps\)
\For{episode \(= 1\) to \(M\)}
    \State \(env, v, d\) \(\leftarrow\) sampleRandomEnv('train', \(T\))
    \State Sample initial state \(s_1\) from \(d_0\) and get \(A_{\text{valid}, 1}\)
    \For{\(t = 1\) to \(N\)}
        \State \(totalSteps \mathrel{{+}{=}} 1\)
        \State Identify \(k\) most relevant actions using Guide:
        \State \(\hat{A}_{\text{relevant}, t} \leftarrow \text{Guide.top\_k}(A_{\text{valid}, t}, k, d_{T, v})\)
        \State \(randomNumber \sim \text{Uniform}(0, 1)\)
        \If{\(randomNumber > \epsilon\)}
            \State \(a_t \sim \text{Multinomial}(\text{softmax}(\{Q(s_t, a | \theta) \text{ for } a \in \hat{A}_{\text{relevant}, t}\}))\)
        \Else
            \State \(a_t \sim \text{Multinomial}(\text{softmax}(\{Q(s_t, a | \theta) \text{ for } a \in A_{\text{valid}, t}\}))\)
        \EndIf
        \State Execute \(a_t\), observe \(r_{t+1}\), \(s_{t+1}\), \(A_{\text{valid}, t+1}\)
        \State Store \((s_t, a_t, r_{t+1}, s_{t+1}, A_{\text{valid}, t+1})\) in \(D\)
        \If{\(totalSteps \mod updateFrequency = 0\)}
            \State Sample batch from \(D\)
            \State \(L_{\text{cumulative}} = 0\)
            \For{each \((s, a, r, s', A')\) in batch}
                \State \(\delta = r + \gamma \max_{a' \in A'} Q(s', a'| \theta) - Q(s, a| \theta)\)
                \State Compute Huber loss \(L\):
                \State \(L = \begin{cases} 
                  \frac{1}{2} \delta^2 & \text{if } |\delta| < 1 \\
                  |\delta| - \frac{1}{2} & \text{otherwise}
               \end{cases}\)
                \State \(L_{\text{cumulative}} \mathrel{{+}{=}} L\)
            \EndFor
            \State Update \(\theta\) with Adam optimizer:
            \State \(\theta \leftarrow \text{AdamOptimizer}(\theta, \nabla_{\theta} L_{\text{cumulative}})\)
        \EndIf
        \State Update state: \(s_t \leftarrow s_{t+1}\)
    \EndFor
\EndFor
\end{algorithmic}
\end{algorithm*}

%% file: tables/ex2.tex
\begin{table*}[!h]
\centering
\resizebox{\textwidth}{!}{%
\begin{tabular}{|l|l|}
    \toprule
    Relevant Gold Actions & Selected By Pruner  \\
    \toprule
     
    &
    focus on inclined plane with a soapy water surface,\textcolor{darkgreen}{look at inclined plane with a soapy water surface}, \\
    look around, & \textcolor{darkgreen}{move block to inclined plane with a soapy water surface, look at inclined plane with a steel surface,} \\
    move block to inclined plane with a steel surface, & \textcolor{darkgreen}{move block to inclined plane with a steel surface, focus on inclined plane with a steel surface}, \\
    focus on inclined plane with a steel surface,go to hallway, & go to hallway, \textcolor{darkgreen}{look around,wait1}connect red wire terminal 2 to anode in green light bulb, \\
    wait1,look at inclined plane with a soapy water surface, & connect red wire terminal 2 to cathode in green light bulb,  \\
    move block to inclined plane with a soapy water surface, & connect battery cathode to red wire terminal 1, \\
    look at inclined plane with a steel surface & connect black wire terminal 2 to anode in green light bulb,  \\
    & connect red wire terminal 2 to anode in red light bulb,\\
    & connect black wire terminal 2 to cathode in green light bulb,
 \\
    & connect battery cathode to black wire terminal 1, \\ 
    & connect red wire terminal 2 to cathode in red light bulb,  \\
    & connect black wire terminal 2 to anode in red light bulb, \\ 
    &  connect black wire terminal 2 to cathode in red light bulb, open freezer, wait, pick up red wire \\
    & focus on red light bulb,pick up black wire, focus on green light bulb, pick up green light bulb, \\
    & pick up black wire, focus on green light bulb, pick up green light bulb \\ \hline
\end{tabular}
}
\caption{\footnotesize Qualitative analysis of Validation set trajectories for the ScienceWorld Task "Friction Known Surfaces" for variation 0 at step 17. Note: Missed gold actions are in \textcolor{red}{Red}, while selected gold actions are in \textcolor{darkgreen}{Green}}
\label{tab:qualitative-analysis_valid_trajectories_task0} 
\end{table*}